\documentclass{article}

\usepackage{microtype}
\usepackage{graphicx}
\usepackage{subcaption}
\usepackage{booktabs} % for professional tables

\usepackage{hyperref}

\usepackage[accepted]{icml2026}

\usepackage{amsmath}
\usepackage{amssymb}
\usepackage{mathtools}
\usepackage{amsthm}
\usepackage{xspace}
\usepackage{float}  % in preamble

\usepackage[capitalize,noabbrev]{cleveref}

\theoremstyle{plain}

\theoremstyle{definition}

\theoremstyle{remark}

\usepackage[textsize=tiny]{todonotes}

\icmltitlerunning{}

\begin{document}

\newcommand{\sysname}{QuantumQUBO Agent\xspace}

\twocolumn[
 \icmltitle{\sysname: Automating Quadratic Unconstrained Binary Optimization (QUBO) Formulation Generation from Natural Language}



  \icmlsetsymbol{equal}{*}

  \begin{icmlauthorlist}
    \icmlauthor{Niloy Kumar Mondal}{yyy}
    \icmlauthor{Md Rizwan Parvez}{comp}
  \end{icmlauthorlist}

  \icmlaffiliation{yyy}{Bangladesh University of Engineering and Technology}
  \icmlaffiliation{comp}{Qatar Computing Research Institute, HBKU, Doha, Qatar}
  
  \icmlcorrespondingauthor{Niloy Kumar Mondal}{nkm2105044@gmail.com}
  \icmlcorrespondingauthor{Md Rizwan Parvez}{mparvez@hbku.edu.qa}

  \icmlkeywords{Machine Learning, ICML}

  \vskip 0.3in
]

\printAffiliationsAndNotice{}  % no special notice (required even if empty)

\begin{abstract}
Quadratic Unconstrained Binary Optimization (QUBO) is a central formulation for combinatorial optimization and has gained increasing attention due to its compatibility with quantum, hybrid quantum-classical, and quantum-inspired solvers. However, translating natural-language problem descriptions into correct QUBO formulations remains difficult, requiring the identification of binary variables, constraints, objective functions, penalty terms, and suitable penalty weights. This process is time-consuming and often demands substantial domain expertise. To address this challenge, we propose an end-to-end multi-agent framework that automatically generates QUBO formulations from natural-language problem descriptions, supported by structured or unstructured test cases. To evaluate its performance, We also introduce QUBOBench, a benchmark containing 100 combinatorial optimization problems across 12 application domains, curated from peer-reviewed literature, competitions, and canonical NP-hard problems. Experimental results show that our framework achieves 68\%  accuracy on QUBOBench, outperforming a direct single-call baseline by 22\%. Further analysis identifies iterative self-repair as the most important component contributing to improved performance. The data and code are open-sourced at \url{https://quitttcat.github.io/QuantumQUBOAgent/}.
\end{abstract}

\section{Introduction}
The Quadratic Unconstrained Binary Optimization (QUBO) problem is a combinatorial optimization problem classified as NP-hard, meaning no efficient algorithm is known to solve it optimally in polynomial time. It has a variety of application areas such as Computer Science, Finance, Physics, Engineering, Medicine, VLSI design, Quantum Computing, Traffic Management, Machine Scheduling, and many more \cite{punnen2022quadratic}.
With advances in quantum computing \cite{Preskill2018quantumcomputingin}, QUBO problems have become increasingly important because they can be addressed by quantum, hybrid quantum-classical, and quantum-inspired methods such as quantum annealing \cite{rajak2023quantum} and the Quantum Approximate Optimization Algorithm (QAOA) \cite{farhi2016quantum}.
Quantum annealing is a powerful heuristic for solving optimization problems \cite{somma2012quantum}.

Researchers across many domains are exploring quantum optimization algorithms with the goal of potential speedups and improved accuracy. These algorithms have been applied to machine learning tasks such as feature selection \cite{mucke2023feature, samanta2025parametric, pathak2025feature} and Bayesian network structure learning \cite{shikuri2025decomposed}, computer vision problems including image segmentation and denoising \cite{heidari2024quantum, kerger2023quantum, venkatesh2024q}, supply chain \cite{moncayo2025quantum, ciacco2026quantum}, drug discovery \cite{yang2025ligand}, RNA folding \cite{fox2022rna}, satellite observation \cite{ferrari2025qubo}, scheduling problems \cite{zhou2025hybrid, gebler2024qubo}, cybersecurity \cite{marini2025cyber}, IoT \cite{zhang2025hybrid}, and computational chemistry \cite{negre2022toward}.

To apply quantum algorithms across domains, optimization problems must first be reformulated as QUBO and often as Ising formulations \cite{lucas2014ising}. In general, given a natural-language description of a combinatorial optimization problem, the task is to find the values of binary decision variables $x \in \{0,1\}^n$ that minimize a cost function of the form
\begin{equation}
    \min_{x \in \{0,1\}^n} \; E(x) \;=\; x^{\top} Q\, x \;+\; c,
\end{equation}
where $Q$ is a matrix of coefficients and $c$ is a constant offset\cite{glover2019quantum}. The solution space contains $2^n$ candidate bitstrings, and finding the optimum is NP-hard in general. 

The cognitive task therefore involves identifying the binary decision variables, extracting the implicit and explicit constraints, designing the cost function and penalty terms for constraint violations, selecting appropriate penalty weights, and finally solving the resulting QUBO using classical heuristics or state-of-the-art quantum solvers. These steps are extremely challenging and time-consuming, as such a complex task often requires domain expertise, mathematical rigor, and experimental feedback.

To address this cognitive bottleneck, we propose an end-to-end multi-agent framework that generates the QUBO formulation from natural language problem description, together with either structured or unstructured test cases.To evaluate our framework, we introduce QUBOBench, a 100 combinatorial optimization problems 12 application domains, curated from peer-reviewed venues, competitions, and canonical NP-hard problems. Our framework achieves 68\% accuracy on QUBOBench, outperforming the direct single-call baseline (46\%) by a 22 percentage margin, with iterative self repair identified as the most critical component driving performance.

\section{Related Work}

Recent work has studied how large language models can translate natural
language problem descriptions into formal optimization models.
The NL4Opt competition established this direction by asking systems to
formulate optimization problems from descriptions in natural language 
\cite{ramamonjison2023nl4opt}. Subsequent systems expand this idea with more
structured reasoning and agentic workflows. Chain-of-Experts decomposes complex
operations-research problems across multiple expert-style LLMs agents
\cite{xiao2024chain}, while Optimus uses LLMs to generate scalable
mixed-integer and linear programming models for downstream solvers
\cite{ahmaditeshnizi2024optimus}. OptimAI and LLMOPT further explore
LLM-powered agents that define, model, and solve general optimization problems
from natural language \cite{thind2025optimai,shu2025llmopt}.

Prior LLM-based approaches to optimization modeling primarily focus on LP(Linear Programming),
MILP (Mixed-Integer Linear Programming), or general solver-ready formulations, whereas the QUBO setting remains
comparatively less explored despite its central role in quantum and
quantum-inspired optimization. However, AutoQUBO automates QUBO generation from
formal programmatic inputs, but still assumes that the user can provide a
precise computational description of the problem and does not address
formulation from natural-language research descriptions
\cite{moraglio2022autoqubo}.

LLM-QUBO is the most closely related work which
studies the automatic conversion of natural-language optimization
descriptions into QUBO formulation. However, its experiments focus mainly on
 MILP to QUBO conversion and validate correctness through structural checks such
as input-format adherence, binarization, slack-variable introduction, and
penalty-term validity \cite{zhang2025llm}. Thus, it provides limited support
for end-to-end validation of the generated formulation as an executable
research artifact. Besides, it does not systematically verify that the natural-language
specification, structured formulation, generated implementation, and
small-instance optima are mutually consistent, nor does it provide a
feedback-driven reformulation process when errors occur. In addition, the empirical evaluation is confined to a small number of canonical textbook problems and does not assess performance across diverse, domain-specific optimization tasks.

\section{Methodology}
\label{sec:methodology}

\begin{figure*}[t]
    \centering
    \includegraphics[width=\textwidth]{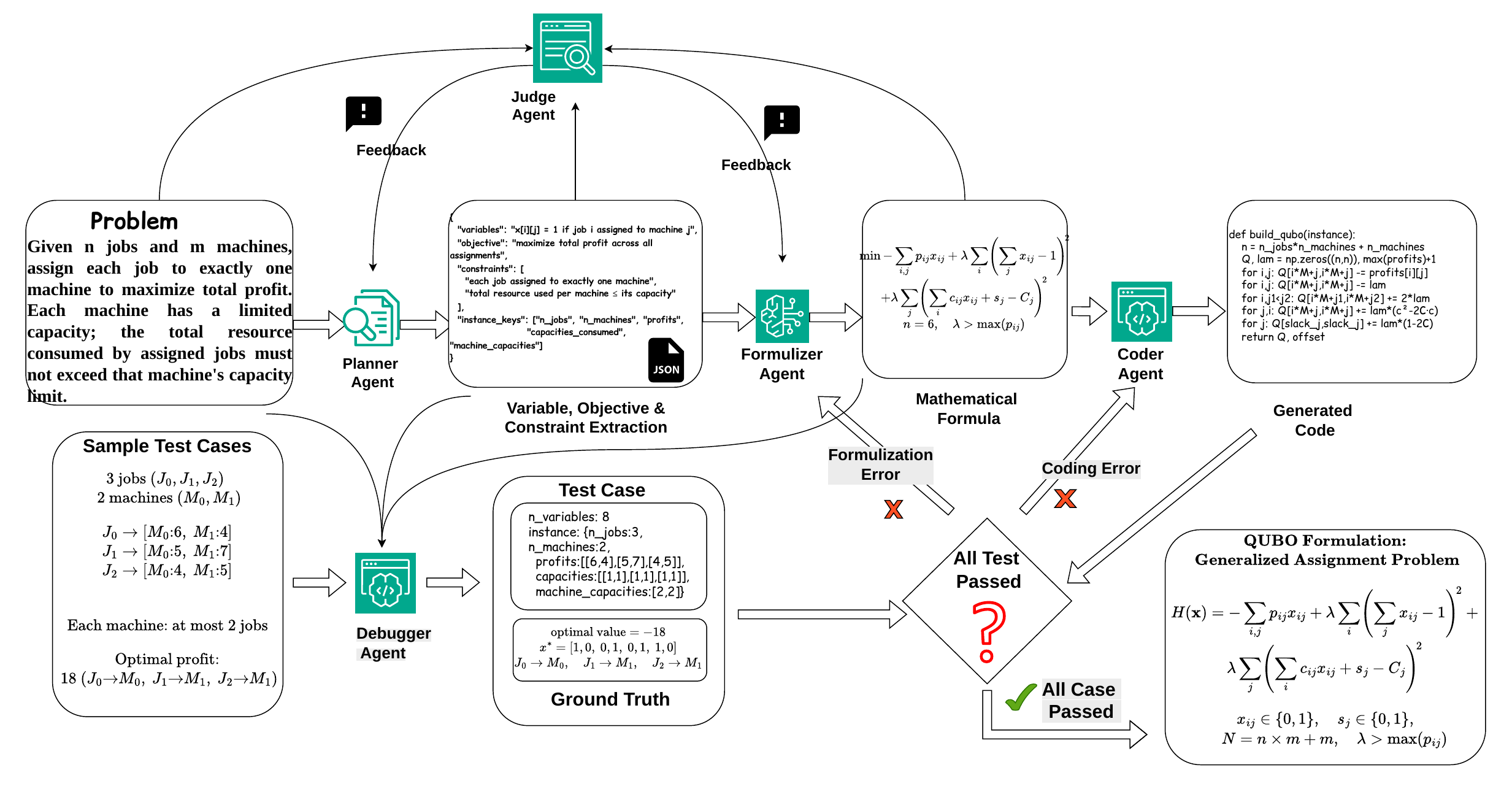}
    \caption{Workflow and Architecture of \sysname}
    \label{fig:architecture}
\end{figure*}

Generating a QUBO formulation from a natural-language problem statement involves several distinct steps: understanding the problem, choosing binary encoding variables and constraints, defining penalty terms, deriving the cost function, generating runnable code, and verifying correctness. Rather than asking a single LLM to handle all of these steps at once, we decompose the task across specialized agents, each responsible for one step. The overall architecture is illustrated in Figure~\ref{fig:architecture}. The role of each agent and the end-to-end workflow are described in the following subsections.

\subsection{Multi-Agent Architecture}
Our framework consists of six LLM agents: Planner Agent, Formulizer Agent, Coder Agent, Debugger Agent, Writer Agent and a Judge Agent.

\subsubsection{The Planner Agent}
Our first agent is the Planner Agent.
It is an LLM agent that reads the natural-language problem description provided by the user and extracts the decision variables, the optimization direction, and every implicit or explicit constraint together with its type (equality, $\leq$, or $\geq$) and hardness. It then returns a structured JSON representation of the problem, which is passed to the Judge Agent for consistency checking against the original description. If the Judge identifies any inconsistency, it returns a list of issues to the Planner, forming a feedback loop. The Planner then performs another round of parsing and planning from the original description guided by the issue list.

\subsubsection{The Formulizer Agent}

Our second agent is the Formulizer Agent.
The Formulizer Agent is an LLM agent that takes the structured JSON representation from the Planner and derives the corresponding QUBO cost function from the decision variables, constraints, and the type of optimization, and simplifies the cost function. The returned formulation, in structured JSON containing the variable mapping, objective and penalty terms, penalty-weight justification, and constant offset, is then passed to the Judge Agent; if rejected, the Formulizer regenerates the formulation guided by the reported issues. The Formulizer takes feedback from the Judge Agent as well as from failed test cases at the end of the pipeline, and tries to improve the formulation.

\subsubsection{The Judge Agent}

Our next agent is the Judge Agent.
It is an LLM agent invoked at two checkpoints: it first checks the Planner's structured specification for consistency with the original
description, and later checks the Formulizer's QUBO cost for consistency with
the specification judging correctness constraint encodings, penalty forms,
and objective sign. In both roles it returns a pass/fail verdict with a list
of specific issues, which are looped back to the originating agent as
feedback for a targeted retry.

\subsubsection{The Debugger Agent}
Our next agent is the Debugger Agent.
It is an LLM agent that is essentially a parser. It converts the user supplied test cases and ground truth from natural language into
structured JSON records suitable for automated verification. It uses the
variable mapping passed from the structured JSON formulation to encode the ground optimal selections as bitstrings. The Debugger never sees the Coder's output or the generated cost encoding matrix, which ensures optima from the generated formula is never stored as ground truth.Besides, the coder never sees the ground truth as well.

\subsubsection{The Coder Agent}
The Coder Agent is an LLM agent that translates the  QUBO formulation into executable Python code. It is
given the structured specification,formulation, and the parameters of a single sample instance but not the ground-truth bitstrings which prevents information leak. The generated function is then handed to the Test Runner for empirical verification, and any failure message is returned to the Coder as feedback for the next attempt.

\subsubsection{The Test Runner}
The Test Runner is a deterministic component (not an LLM agent) that validates the Coder's output against the ground truth on each test case. It classifies any failure as either a \emph{coding error} (an exception or a malformed function) or a \emph{formulation error} (a valid output whose optimum disagrees with the ground truth). Coding errors are routed back to the Coder Agent, while formulation errors are routed back to the Formulizer Agent for a fresh derivation guided by the reported issues.

\subsubsection{The Writer Agent}
Our final agent is the writer agent that produces a visual representation and summary of the output.

\section{QUBOBench}
To evaluate our framework on domain specific problems, we introduce QUBOBench, a benchmark comprising 100 combinatorial optimization problems and 200 small instance test cases spanning 12 application domains. The problem instances are curated from three complementary sources: 76 drawn from peer-reviewed venues, 5 from competitions and hackathons, and 19 canonical NP-hard problems primarily adapted from Lucas~\cite{lucas2014ising}. A data sample consists of hand curated problem description and sample small instance test cases both in natural language. For more details, refer to Appendix~\ref{app:dataset}.
\section{Experimental Setup}

\subsection{Evaluation Protocal}

We evaluate \sysname{} on the QUBOBench dataset. Performance is measured across two metrics:
\textit{accuracy}, defined as the fraction of benchmarks for which the pipeline
produces a QUBO formulation and matches all ground-truth optimal
solutions;
and \textit{total token consumption}.

\subsection{Baseline Metric}

As a lower-bound baseline, we conduct a direct evaluation in which the
natural-language problem description is presented directly to the LLM with
a single prompt. To ensure fair and fully
automated evaluation, we provide the model with structured instance dictionaries
previously generated by the Debugger agent, constraining the output to a fixed function signature. This eliminates the
possibility of format-induced false negatives in automated scoring. Ground-truth
optimal solutions are never exposed to the model at any stage. 

\subsection{Model Selection}
All reasoning agents (Planner, Judge, Formulizer, Debugger) use \texttt{qwen/qwen3-235b-a22b}, while the Coder agent uses \texttt{qwen/qwen3-coder-next}.
We believe our pipeline is LLM-agnostic.
For the direct evaluation we used \texttt{qwen/qwen3-coder-next}.

\subsection{Ablation Study}
We conduct two ablation studies: removing the iterative feedback once and removing the judge together with feedback, to isolate the contribution.

\section{Results and Discussions}

\begin{table}[H]
  \begin{center}
    \begin{small}
      \begin{sc}
        \resizebox{\columnwidth}{!}{%
        \begin{tabular}{ccccccc}
          \toprule
          Planner & Judge & Formulizer & Coder & Iter.\ Feedback & Acc.\ (\%) & Tokens (k) \\
          \midrule
          $\checkmark$ & $\checkmark$ & $\checkmark$ & $\checkmark$ & $\checkmark$ & \textbf{68} & 48.1 \\
          $\checkmark$ & $\checkmark$ & $\checkmark$ & $\checkmark$ & $\times$     & 35          & 24.7 \\
          $\checkmark$ & $\times$     & $\checkmark$ & $\checkmark$ & $\times$     & 36          & 22.1 \\
          $\times$     & $\times$     & $\times$     & $\times$     & $\times$     & 46          &  3.2 \\
          \bottomrule
        \end{tabular}}
      \end{sc}
    \end{small}
  \end{center}
  \caption{Ablation results on QUBOBench. Rows from top: \sysname, without iterative feedback, without Judge and iterative feedback, Direct Method.}
  \label{tab:main}
  \vskip -0.1in
\end{table}

Table~\ref{tab:main} shows that the full pipeline achieves 68\% accuracy, 
outperforming all ablated variants. Removing iterative feedback 
causes the largest drop to 35\%, confirming that self-repair is the most 
critical component. Removing the Judge while keeping all other components 
fixed yields 36\%, nearly identical to the no-retry setting, indicating that 
the Judge provides value only when paired with a sufficient retry budget to 
act on its feedback.

\subsection{Failure mode analysis}

A \textit{formulation error} occurs when 
the agent produces an incorrect QUBO objective, wrong penalty 
terms, incorrect variable mapping, or an energy landscape that 
does not encode the problem correctly. A \textit{coding error} 
occurs when the QUBO mathematics is correct but the 
\texttt{build\_qubo} implementation fails due to index bugs, 
wrong array shapes, or incorrect NumPy operations.

\begin{table}[H]
  \begin{center}
    \begin{small}
      \begin{sc}
        \resizebox{\columnwidth}{!}{%
        \begin{tabular}{lccc}
          \toprule
          Method & Code Error (\%) & Form.\ Error (\%) & Other (\%) \\
          \midrule
          QuantumQUBO Agent & 79.3 & 15.5 &  5.2 \\
          Direct Method     & 26.4 & 66.0 &  7.5 \\
          \bottomrule
        \end{tabular}}
      \end{sc}
    \end{small}
  \end{center}
  \caption{Error type distribution on QUBOBench. Percentages are of total errors by \sysname vs direct method.}
  \label{tab:failure}
  \vskip -0.1in
\end{table}

In the \sysname, 79.3\% of all errors originate
from the Coder agent, while only 15.5\%  from formulation 
failures, indicating that the Planner-Formulizer-Judge
chain successfully resolves most mathematical ambiguities.
In contrast, the direct method shows the inverse pattern:
66.0\% of failures are formulation errors, confirming that
without structured reasoning, a single LLM call struggles
primarily with deriving correct QUBO mathematics rather
than implementing it.

\section{Future Work}
Future work includes fine-tuning separate agents for specific tasks, adding Ising Formulator incorporated with quantum hardware feedback.

\bibliography{example_paper}
\bibliographystyle{icml2026}

% %%%%%%%%%%%%%%%%%%%%%%%%%%%%%%%%%%%%%%%%%%%%%%%%%%%%%%%%%%%%%%%%%%%%%%%%%%%%%%%
% %%%%%%%%%%%%%%%%%%%%%%%%%%%%%%%%%%%%%%%%%%%%%%%%%%%%%%%%%%%%%%%%%%%%%%%%%%%%%%%
% APPENDIX
%%%%%%%%%%%%%%%%%%%%%%%%%%%%%%%%%%%%%%%%%%%%%%%%%%%%%%%%%%%%%%%%%%%%%%%%%%%%%%%
%%%%%%%%%%%%%%%%%%%%%%%%%%%%%%%%%%%%%%%%%%%%%%%%%%%%%%%%%%%%%%%%%%%%%%%%%%%%%%%
\newpage
\appendix
\onecolumn

\section{Algorithm of \sysname}

\textbf{Input:} A natural-language problem description $P$, a set of test cases $\mathcal{T}$,
retry budgets $R_{\text{planner}}, R_{\text{formulizer}}, R_{\text{coder}}$ controlling the maximum
number of repair attempts at each stage, and a pass threshold $\tau \in (0, 1]$ defining
the minimum fraction of test cases that must pass for a solution to be accepted.

\textbf{Output:} A verified QUBO formulation $\mathcal{F}$, a correctness-verified builder
function $f$ that constructs the QUBO matrix for any problem instance, a worked example,
and a \LaTeX{} write-up with compiled PDF deliverable.

\begin{algorithm}[H]
\caption{QuantumQUBO Agent Pipeline}
\label{alg:pipeline}
\begin{algorithmic}[1]
  \STATE $R_{\text{planner}}, R_{\text{formulizer}}, R_{\text{coder}} \leftarrow$ number of maximum retry for each stage
  \STATE $\tau \leftarrow$ minimum pass rate to accept a solution
  \STATE
  \STATE $\mathit{spec} \leftarrow \text{PlannerAgent}(P)$
  \STATE $\mathit{check} \leftarrow \text{JudgeAgent}(P,\ \mathit{spec})$
  \FOR{$i \leftarrow 1$ \textbf{to} $R_{\text{planner}}$}
    \IF{\textbf{not} $\mathit{check}.\mathit{passed}$}
      \STATE $\mathit{spec} \leftarrow \text{PlannerAgent}(P,\ \mathit{check}.\mathit{issues})$
      \STATE $\mathit{check} \leftarrow \text{JudgeAgent}(P,\ \mathit{spec})$
    \ENDIF
  \ENDFOR
  \IF{\textbf{not} $\mathit{check}.\mathit{passed}$} \STATE \textbf{return} Fail \ENDIF
  \STATE
  \STATE $\mathit{formulation} \leftarrow \text{FormulizationAgent}(\mathit{spec})$
  \STATE $\mathit{check} \leftarrow \text{JudgeAgent}(\mathit{spec},\ \mathit{formulation})$
  \FOR{$i \leftarrow 1$ \textbf{to} $R_{\text{formulizer}}$}
    \IF{\textbf{not} $\mathit{check}.\mathit{passed}$}
      \STATE $\mathit{formulation} \leftarrow \text{FormulizationAgent}(\mathit{spec},\ \mathit{check}.\mathit{issues})$
      \STATE $\mathit{check} \leftarrow \text{JudgeAgent}(\mathit{spec},\ \mathit{formulation})$
    \ENDIF
  \ENDFOR
  \IF{\textbf{not} $\mathit{check}.\mathit{passed}$} \STATE \textbf{return} Fail \ENDIF
  \STATE
  \STATE $\mathit{test\_cases} \leftarrow \text{DebuggerAgent}(\mathit{spec},\ \mathit{mapping(formulation)},\ \mathit{sample\_input})$
  \STATE
\FOR{$i \leftarrow 1$ \textbf{to} $R_{\text{coder}}$}
    \STATE $\mathit{code} \leftarrow \text{CoderAgent}(\mathit{spec},\ \mathit{formulation},\ \mathit{error})$
    \STATE $\mathit{results},\ \mathit{class} \leftarrow \text{TestRunner}(\mathit{code},\ \mathit{test\_cases})$
    \STATE $\mathit{pass\_rate} \leftarrow \mathit{results}.\mathit{passed}\ /\ |\mathit{test\_cases}|$
    \IF{$\mathit{pass\_rate} \geq \tau$}
      \STATE $\mathit{latex} \leftarrow \text{WriterAgent}(\mathit{spec},\ \mathit{formulation})$
      \STATE \textbf{return} $\mathit{formulation},\ \mathit{code},\ \mathit{latex}$
    \ENDIF
    \IF{$\mathit{class} = \textit{formulation\_error}$}
      \STATE $\mathit{formulation} \leftarrow \text{FormulizationAgent}(\mathit{spec},\ \mathit{results}.\mathit{errors})$
      \STATE $\mathit{check} \leftarrow \text{JudgeAgent}(\mathit{spec},\ \mathit{formulation})$
    \ENDIF
    \STATE $\mathit{error} \leftarrow \mathit{results}.\mathit{errors}$
  \ENDFOR
  \STATE \textbf{return} Fail
\end{algorithmic}
\end{algorithm}

\section{Details of Dataset}

\label{app:dataset}

\subsection{Overview}

QUBOBench is a benchmark designed to evaluate Large Language Models (LLMs) on the task of formulating Quadratic Unconstrained Binary Optimization (QUBO) problems from natural-language descriptions. The dataset comprises 100 problem instances and 200 test cases spanning 12 application domains, encompassing classical graph-theoretic and combinatorial problems (e.g., Max-Cut, Maximum Independent Set, Graph Coloring, Hamiltonian Path), logistics and scheduling (Vehicle Routing, Job Shop Scheduling, Bin Packing), machine learning tasks formulated as discrete optimization (feature selection, $k$-means and $k$-medoids clustering, Bayesian network structure learning), computational biology and chemistry (RNA folding, protein lattice folding, molecular conformation selection), finance (portfolio optimization, credit scorecard selection), computer vision (image segmentation, denoising, stereo matching), networks and infrastructure (sensor placement, distribution-network reconfiguration), and quantum computing itself (qubit allocation, circuit compilation, prime factorization).

\subsection{Dataset Structure}

Each datapoint in QUBOBench consists of two files:
\begin{itemize}
    \item \texttt{prompt.txt} : a natural-language description of the optimization problem, expressed in the style and terminology of its source domain.
    \item \texttt{sample\_cases.txt} : one or more natural-language test cases that specify a concrete problem instance, the optimal solution configuration, and the corresponding ground-truth objective value. Each test case explicitly cites its source.
\end{itemize}
This format preserves the natural-language character of the input, ensuring that the benchmark evaluates an LLM's ability to parse domain-specific terminology, identify decision variables and constraints, and derive a valid QUBO formulation without relying on pre-structured inputs.

\subsection{Problem Sources}

The 100 problem instances are curated from three complementary sources to balance rigor, diversity, and coverage.

\paragraph{(1) Peer-reviewed research papers (76 problems).} The majority of instances are drawn from journal articles and conference papers with verifiable DOIs. Contributing venues include:

\textit{Journals} --- \emph{The Journal of Finance}, \emph{Management Science}, \emph{Artificial Intelligence}, \emph{PLOS Computational Biology}, \emph{PLOS ONE}, \emph{Nature Scientific Reports}, \emph{npj Quantum Information}, \emph{npj Unconventional Computing}, \emph{Physical Review Research}, \emph{Journal of Chemical Theory and Computation}, \emph{Theoretical Computer Science}, \emph{Quantum Machine Intelligence}, \emph{Quantum Information Processing}, \emph{Annals of Operations Research}, \emph{Journal of Combinatorial Optimization}, \emph{Journal of Optimization Theory and Applications}, \emph{Applied Soft Computing}, \emph{Future Generation Computer Systems}, \emph{Neurocomputing}, \emph{Electric Power Systems Research}, \emph{Discrete Optimization}, \emph{IEEE Transactions on Quantum Engineering}, \emph{IEEE Internet of Things Journal}, \emph{IEEE Computer Graphics and Applications}, \emph{Entropy}, \emph{Applied Sciences}, \emph{Journal of Risk and Financial Management}, \emph{Frontiers in Physics}, \emph{Frontiers in Computer Science}, \emph{Frontiers in ICT}, \emph{Bioinformatics Advances}, \emph{International Journal for Numerical Methods in Engineering}, \emph{EPJ Quantum Technology}, \emph{Results in Engineering}, \emph{Array}, \emph{SN Computer Science}, \emph{Journal of Membrane Computing}, \emph{IEICE Communications Express}, and \emph{Journal of Information Processing}.

\textit{Conferences} --- \emph{AAAI}, \emph{ACM SIGKDD}, \emph{ACM/SIGDA FPGA}, \emph{IEEE BigData}, \emph{IEEE GLOBECOM}, \emph{IEEE QAI}, \emph{IEEE QCE}, \emph{IEEE EIT}, \emph{ICCSA}, \emph{JSME Robotics and Mechatronics}, and \emph{IFAC}.

\paragraph{(2) Competitions and hackathons (5 problems).} To capture practitioner-style framings and applied challenges, we include problems posed in:
\begin{itemize}
    \item Fixstars Amplify Benchmark Suite
    \item Aqora / U.S.\ DOE Global Industry Challenge 2026
    \item iQuHACK 2025 D-Wave Hackathon
    \item AtCoder Educational DP Contest
    \item D-Wave Examples Repository
\end{itemize}

\paragraph{(3) Canonical NP-hard problems (19 problems).} Classical decision and optimization problems including Max-Cut, Hamiltonian Cycle, Travelling Salesman, Subset Sum, and Vertex Cover which are drawn primarily from Lucas~\cite{lucas2014ising}, supplemented with hand-curated textbook instances. These problems serve as foundational test cases against which any general-purpose QUBO-formulation system should perform reliably.

All natural-language prompts and test cases were hand-curated by the authors to ensure clarity, consistency, and faithful representation of each source problem.

\subsection{Domain-wise Distribution}

\begin{figure}[h]
\centering
\begin{minipage}{0.45\textwidth}
\centering
\captionof{table}{Per-domain distribution of problems and test cases in QUBOBench.}
\label{tab:domain-dist}
\begin{tabular}{lcc}
\toprule
Domain & \#Problems & \#TC \\
\midrule
Graph Theory & 23 & 49 \\
Scheduling & 11 & 25 \\
Machine Learning & 11 & 22 \\
Logistics \& Transportation & 12 & 21 \\
Biology \& Chemistry & 9 & 19 \\
Networks \& Infrastructure & 7 & 15 \\
Combinatorics \& Logic & 8 & 11 \\
Finance & 5 & 10 \\
Computer Vision & 5 & 10 \\
Quantum Computing & 5 & 10 \\
Robotics & 2 & 4 \\
Electronics \& VLSI & 2 & 4 \\
\midrule
\textbf{Total} & \textbf{100} & \textbf{200} \\
\bottomrule
\end{tabular}
\end{minipage}
\hfill
\begin{minipage}{0.5\textwidth}
\centering
\includegraphics[width=\linewidth]{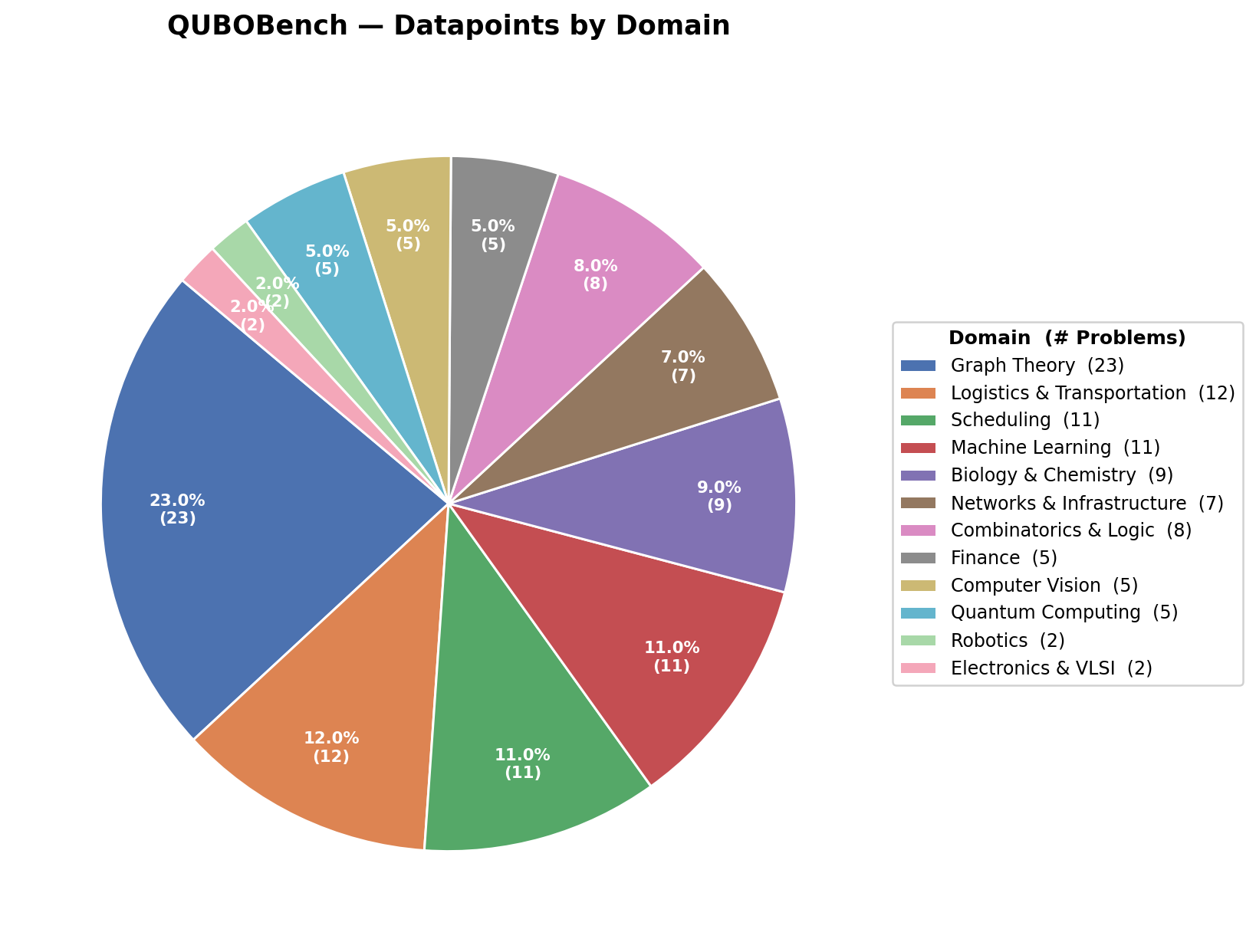}
\captionof{figure}{Per-domain distribution of problems in QUBOBench.}
\label{fig:domain-pie}
\end{minipage}
\end{figure}

Table~\ref{tab:domain-dist} summarizes the per-domain distribution of problems and test cases. Graph theory dominates due to the centrality of graph-theoretic formulations in the QUBO literature, while domains such as Robotics and Electronics \& VLSI remain comparatively sparse, reflecting the relative scarcity of published QUBO formulations in these areas.

% You can have as much text here as you want. The main body must be at most $8$
% pages long. For the final version, one more page can be added. If you want, you
% can use an appendix like this one.

% The $\mathtt{\backslash onecolumn}$ command above can be kept in place if you
% prefer a one-column appendix, or can be removed if you prefer a two-column
% appendix.  Apart from this possible change, the style (font size, spacing,
% margins, page numbering, etc.) should be kept the same as the main body.
% %%%%%%%%%%%%%%%%%%%%%%%%%%%%%%%%%%%%%%%%%%%%%%%%%%%%%%%%%%%%%%%%%%%%%%%%%%%%%%%
%%%%%%%%%%%%%%%%%%%%%%%%%%%%%%%%%%%%%%%%%%%%%%%%%%%%%%%%%%%%%%%%%%%%%%%%%%%%%%%

\end{document}